\documentclass{article} 
\usepackage{iclr2017_conference,times}
\usepackage{hyperref}
\usepackage{url}
\usepackage{times}
\usepackage{helvet}
\usepackage{courier}
\usepackage{graphicx}
\usepackage{algorithm}
\usepackage{multirow}
\usepackage{algorithmic}

\usepackage{amsmath}
\usepackage{amsfonts}
\title{Can AI Generate Love Advice?:\\
 Toward Neural Answer Generation \\ for Non-Factoid Questions}

\author{Makoto Nakatsuji \\
NTT Resonant Inc.\\
 \\
\texttt{nakatsuji.makoto@gmail.com} \\
}

\begin{document}

\maketitle

\begin{abstract}
   Deep learning methods that extract answers for non-factoid questions
   from QA sites are seen as critical since they can assist users in
   reaching their next decisions through conversations with AI systems.
   The current methods, however, have the following two problems: (1)
   They can not understand the ambiguous use of words in the questions
   as word usage can strongly depend on the context (e.g. the word
   ``relationship'' has quite different meanings in the categories of
   Love advice and other categories). As a result, the accuracies of
   their answer selections are not good enough.  (2) The current methods
   can only {\em select} from among the answers held by QA sites and can
   not {\em generate} new ones.  Thus, they can not answer the
   questions that are somewhat different with those stored in QA sites.
    Our solution, Neural Answer Construction Model, tackles these
   problems as it:
   (1) Incorporates the biases of semantics behind questions
   (e.g. categories assigned to questions) into word embeddings while
   also computing them regardless of the semantics.  As a result, it can
   extract answers that suit the contexts of words used in the question
   as well as following the common usage of words across semantics. This improves
   the accuracy of answer selection.  (2) Uses biLSTM to compute the embeddings of
   questions as well as those of the sentences often used to form
   answers (e.g. sentences representing conclusions or those
   supplementing the conclusions). It then simultaneously
   learns the optimum combination of those sentences as well as the
   closeness between the question and those sentences.  As a result, our
   model can construct an answer that corresponds to the situation
   that underlies the question; it fills the gap between answer selection
   and generation and is the first model to move beyond the current
   simple answer selection model for non-factoid QAs.  Evaluations using
   datasets created for love advice stored in the Japanese QA site,
   Oshiete goo, indicate that our model achieves 20 \% higher accuracy
   in answer creation than the strong baselines.  Our model is
   practical and has already been applied to the love advice service in
   Oshiete goo.
\end{abstract}

\section{Introduction}

Recently, dialog-based natural language understanding systems such as
Apple's Siri, IBM's Watson, Amazon's Echo, and Wolfram Alpha have spread
through the market. In those systems, Question Answering (QA) modules
are particularly important since people want to know many things in
their daily lives.
Technically, there are two types of questions in QA systems: factoid
questions and non-factoid ones. The former are asking, for instance, for
the name of a person or a location such that ``What/Who is $X$?''.  The
latter are more diverse questions which cannot be answered by a short
fact. They range from advice on making long distance relationships work
well, to requests for opinions on some public issues.  Significant
progress has been made at answering factoid questions
(\cite{wang-smith-mitamura:2007:EMNLP-CoNLL2007,DBLP:journals/corr/YuHBP14}),
however, retrieving answers for non-factoid questions from the Web
remains a critical challenge in improving QA modules. The QA community
sites such as Yahoo! Answers and Quora can be sources of training data
for the non-factoid questions where the goal is to automatically select
the best of the stored candidate answers.


Recent deep learning methods have been applied to this non-factoid
answer selection task using datasets stored in the QA sites resulting in
state-of-the-art performance
(\cite{DBLP:journals/corr/YuHBP14,TanXZ15,Qiu,DBLP:journals/corr/FengXGWZ15,DBLP:conf/acl/WangN15,TanSXZ16}).
They usually compute closeness between questions and answers by the
individual embeddings obtained using a convolutional model. For example,
\cite{TanSXZ16} builds the embeddings of questions and those of answers
based on bidirectional long short-term memory (biLSTM) models, and
measures their closeness by cosine similarity. It also utilizes an
efficient attention mechanism to generate the answer representation
following the question context. Their results show that their model can
achieve much more accurate results than the strong baseline
(\cite{DBLP:journals/corr/FengXGWZ15}).  The current methods, however,
have the following two problems when applying them to real applications:

(1) They can not understand the ambiguous use of words written in the
questions as words are used in quite different ways following the
context in which they appear (e.g. the word ``relationship'' used in a
question submitted to ``Love advice'' category is quite different from
the same word submitted to ``Business advice'' category). This makes
words important for a specific context likely to be disregarded in the
following answer selection process. As a result, the answer selection
accuracies become weak for real applications.

(2) They can only {\em select} from among the answers stored in the QA
systems and can not {\em generate} new ones.  Thus, they can not answer
the questions that are somewhat different from those stored in the QA
systems even though it is important to cope with such differences when
answering non-factoid questions (e.g. questions in the ``Love advice''
category are often different due to the situation and user even though
they share the same topics.).  Furthermore, the answers selected from QA
datasets often contain a large amount of unrelated information.  Some
other studies have tried to create short answers to the short questions
often seen in chat systems
(\cite{DBLP:journals/corr/VinyalsL15,DBLP:journals/corr/SerbanSBCP15}).
Our target, non-factoid questions in QA systems, are, however, much
longer and more complicated than those in chat systems. As described in
their papers, the above methods, unfortunately, create unsatisfying
answers to such non-factoid questions.

\begin{figure}[t]
\begin{center}
\includegraphics[width=0.8\linewidth]{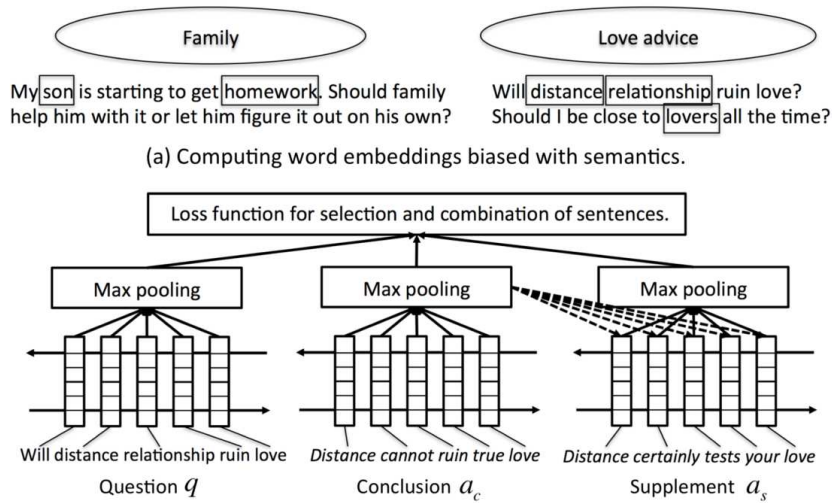}
\end{center}
\vspace{-5mm}
\caption{Main ideas: (a) word embeddings with semantics and (b)
 a neural answer construction.}
 \label{fig:1}
\vspace{-4mm}
\end{figure}


To solve the above problems, this paper proposes a neural answer
construction model; it fills the gap between answer selection and
generation and is the first model to move beyond the current simple
answer selection model for non-factoid QAs.  It extends the above
mentioned biLSTM model since it is language independent and free from
feature engineering, linguistic tools, or external resources. Our model
takes the following two ideas:

(1) Before learning answer creation, it incorporates semantic biases
behind questions (e.g. titles or categories assigned to questions) into
word vectors while computing vectors by using QA documents stored across
semantics.  This process emphasizes the words that are important for a
certain context.  As a result, it can select the answers that suit the
contexts of words used in the questions as well as the common usage of
words seen across semantics. This improves the accuracies of answer
selections. For example, in Fig. \ref{fig:1}-(a), there are two
questions in category ``Family'' and ``Love advice''.  Words marked with
rectangles are category specific (i.e. ``son'' and
``homework'' are specifically observed in ``Family'' while ``distance'',
``relationship'', and ``lovers'' are found in ``Love advice''.) Our method can
emphasize those words. As a result, answers that include the topics,
``son'' and ``homework'', or topics, ``distance'', ``relationship'', and
``lovers'', will be scored highly for the above questions in the
following answer selection task.

(2) The QA module designer first defines the abstract scenario of answer
to be created; types of sentences that should compose the answer and
their occurrence order in the answer (e.g. typical answers in ``Love
advice'' are composed in the order of the sentence types ``sympathy'',
``conclusion'', ``supplementary for conclusion'', and
``encouragement'').  The sentence candidates can be extracted from the
whole answers by applying sentence extraction methods or sentence type
classifiers
(\cite{Schmidt:2014:DST:2637748.2638409,Zhang:2008:STB:1599081.1599218,Nishikawa:2010:OSI:1944566.1944671,Chen:2010:OEP:2898607.2898768}).
It next simultaneously learns {\em the closeness} between questions and
sentences that may include answers as well as {\em combinational
optimization} of those sentences.  Our method also uses an attention
mechanism to generate sentence representations according to the prior
sentence; this extracts important topics in the sentence and tracks
those topics in subsequent sentences.
As a result, it can construct answers that have natural sentence flow
 whose topics correspond to the questions.
Fig. \ref{fig:1}-(b) explains the proposed neural-network by using
examples.  Here, the QA module designer first defines the abstract
scenario for the answer as in the order of ``conclusion'' and
``supplement''.  Thus, there are three types of inputs ``question'',
``conclusion'', and ``supplement''. It next runs biLSTMs over those
inputs separately; it learns the order of word vectors such that
``relationships'' often appears next to ``distance''.  It then computes
the embedding for the question, that for conclusion, and that for
supplement by max-pooling over the hidden vectors output by
biLSTMs. Finally, it computes the closeness between question and
conclusion, that between question and supplement, and combinational
optimization between conclusion and supplement with the attention
mechanism, simultaneously (dotted lines in Fig. \ref{fig:1}-(b)
represent attention from conclusion to supplement).

%


We evaluated our method using datasets stored in the Japanese QA site
Oshiete goo\footnote{http://oshiete.goo.ne.jp}. In particular, our
evaluations focus on questions stored in the ``Love advice'' category
since they are representative non-factoid questions: the questions are
often complicated and most questions are very long.
The results show that our method outperforms the previous methods
including the method by (\cite{TanSXZ16}); our method accurately
constructs answers by naturally combining key sentences that are highly
close to the question.


\section{Related work}


Previous works on answer selection normally require feature engineering,
linguistic tools, or external resources. Recent deep learning methods
are attractive since they demonstrate superior performance compared to
traditional machine learning methods without the above mentioned
tiresome procedures.  For example,
(\cite{DBLP:conf/acl/WangN15,NIPS2014_5550}) construct a joint feature
vector on both question and answer and then convert the task into a
classification or ranking problem.
(\cite{DBLP:journals/corr/FengXGWZ15,DBLP:journals/corr/YuHBP14,dossantos-EtAl:2015:ACL-IJCNLP,Qiu})
learn the question and answer representations and then match them by
certain similarity metrics.  Recently, \cite{TanSXZ16} took the latter
approach and achieved more accurate results than the current strong
baselines
(\cite{DBLP:journals/corr/FengXGWZ15,Bendersky:2011:PCW:2009916.2009998}).
They, however, can only select answers and not generate them.  Other
than the above, recent neural text generation methods
(\cite{DBLP:journals/corr/SerbanSBCP15,DBLP:journals/corr/VinyalsL15}) can
also intrinsically be used for answer generation.  Their evaluations
showed that they could generate very short answer for factoid questions,
but not the longer and more complicated answers demanded by non-factoid
questions.
Our Neural Answer Construction Model fills the gap between answer
selection and generation for non-factoid QAs.  It simultaneously learns
{\em the closeness} between questions and sentences that may include
answers as well as {\em combinational optimization} of those sentences.
Since the sentences themselves in the answer are short, they can be
generated by neural conversation models like
(\cite{DBLP:journals/corr/VinyalsL15});

As for word embeddings with semantics, some previous methods use the
semantics behind words by using semantic lexicons such as WordNet and
Freebase
(\cite{DBLP:conf/cikm/XuBBGWLL14,Bollegala:AAAI:2016,faruqui-EtAl:2015:NAACL-HLT,Johansson-Richard2015-8}). They,
however, do not use the semantics behind the question/answer documents;
e.g. document categories. Thus, they can not well catch the contexts in
which the words appear in the QA documents. They also require external
semantic resources other than QA datasets.


\section{Preliminary}

Here, we explain QA-LSTM (\cite{TanXZ15}), the basic discriminative
framework for answer selection based on LSTM, since we base our ideas on
its framework.

We first explain the LSTM and introduce the terminologies used in this
paper. Given input sequence ${\bf{X}} = \{{\bf{x}}(1), {\bf{x}}(2),
\cdots , {\bf{x}}(N)\}$, where ${\bf{x}}(t)$ is $t$-th word vector, 
$t$-th hidden vector ${\bf{h}}(t)$ is updated as:

\vspace{-2mm}
\begin{eqnarray*}
{\bf{i}}_t &=& \sigma({\bf{W}}_i {\bf{x}}(t) + {\bf{U}}_i {\bf{h}}(t-1) +
 {\bf{b}}_i) \\
{\bf{f}}_t &=& \sigma({\bf{W}}_f {\bf{x}}(t) + {\bf{U}}_f {\bf{h}}(t-1) +
 {\bf{b}}_f) \\
{\bf{o}}_t &=& \sigma({\bf{W}}_o {\bf{x}}(t) + {\bf{U}}_o {\bf{h}}(t-1) +
 {\bf{b}}_o) \\
\widetilde{{\bf{c}}}_t &=& \tanh({\bf{W}}_c {\bf{x}}(t) + {\bf{U}}_c {\bf{h}}(t-1) +
 {\bf{b}}_c) \\
{{\bf{c}}}_t &=& {\bf{i}}_t \ast \widetilde{{\bf{c}}}_t + {\bf{f}}_t \ast {{\bf{c}}_{t-1}}\\
{\bf{h}}(t) &=& {\bf{o}}_t \ast \tanh({\bf{c}}_t)
\end{eqnarray*}

There are three gates (input ${\bf{i}}_t$, forget ${\bf{f}}_t$, and
output ${\bf{o}}_t$), and a cell memory vector ${\bf{c}}_t$. $\sigma$ is
the sigmoid function.  ${\bf{W}} \in R^{H \times N}$, ${\bf{U}} \in R^{H
\times H}$, and ${\bf{b}} \in R^{H \times 1}$ are the network parameters
to be learned.  Single-direction LSTMs are weak in that they fail to
make use of the contextual information from the future tokens. BiLSTMs
use both the previous and future context by processing the sequence in
two directions, and generate two sequences of output vectors. The output
for each token is the concatenation of the two vectors from both
directions, i.e. ${\overrightarrow{h(t)}}={\overrightarrow{h(t)}}
\parallel {\overleftarrow{h(t)}}$.

In the QA-LSTM framework, given input pair $(q, a)$ where $q$ is a
question and $a$ is a candidate answer, it first retrieves the word
embeddings (WEs) of both $q$ and $a$. Next, it separately applies a
biLSTM over the two sequences of WEs. Then, it generates fixed-sized
distributed vector representations ${\bf{o}}_q$ for $q$ (or
${\bf{o}}_a$ for $a$) by computing max pooling over all the output
vectors and then concatenating the resulting vectors on both directions
of the biLSTM. Finally, it uses cosine similarity $\cos({\bf{o}}_q,
{\bf{o}}_a)$ to score the input $(q, a)$ pair.

It then defines the training objective as the hinge loss of:

\vspace{-2mm}
\begin{equation*}
{\cal{L}} = \max\{0, M-\cos({\bf{o}}_q,{\bf{o}}_a^{+}) + \cos({\bf{o}}_q,{\bf{o}}_a^{-})\}
\end{equation*}
where ${\bf{o}}_a^{+}$ is an output vector for ground truth answer,
${\bf{o}}_a^{-}$ is that for an incorrect answer randomly chosen from
the entire answer space, and $M$ is a margin. It treats any question
with more than one ground truth as multiple training examples. Finally,
batch normalization is performed on the representations before computing
cosine similarity (\cite{DBLP:conf/icml/IoffeS15}).


\section{Method}

We first explain our word embeddings with semantics.


\subsection{Word embeddings with document semantics}

This process is inspired by paragraph2vec (\cite{LeM14}); an unsupervised
algorithm that learns fixed-length feature representations from
variable-length pieces of texts, such as sentences, paragraphs, and
documents.

First, we explain paragraph2vec model. It averages the paragraph vector
with several word vectors from a paragraph and predicts the following
word in the given context. It trains both word vectors and paragraph
vectors by stochastic gradient descent and backpropagation
(\cite{Rumelhart104451}).  While paragraph vectors are unique among
paragraphs, the word vectors are shared.

\begin{figure}[t]
\begin{center}
\includegraphics[width=0.7\linewidth]{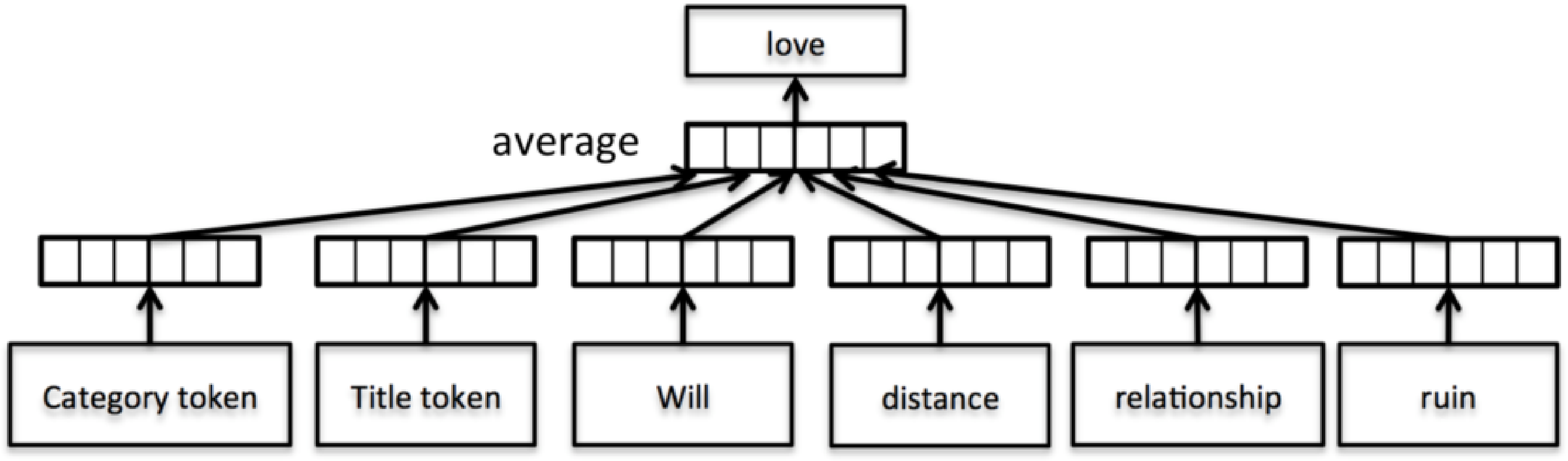}
\end{center}
\vspace{-5mm}
\caption{Learning word vectors biased with semantics.}
 \label{fig:catvec}
\vspace{-4mm}
\end{figure}

Next, we introduce our method that incorporates the semantics behind QA
documents into word embeddings (WEs) in the training phase.  The idea is
simple.  Please see Fig. \ref{fig:catvec}.  It averages the vector
of category token and the vectors of title tokens, which are assigned to
the QA documents, with several of the word vectors present in those
documents. It then predicts the following word in the given context.
Here, title tokens are defined by nouns that are extracted from titles
assigned to the question. Multiple title tokens can be extracted from a
title while one category token is assigned to a question.  Those tokens
are shared among datasets in the same category.  It trains the category
vector and title vectors as well as word vectors in QA documents as per
paragraph2vec model.  Those additional vectors are used as semantic
biases for learning WEs.  They are useful in emphasizing the words
following the contexts of particular categories or titles. This improves
the accuracies of answer selection described later as explained in
Introduction.

For example, in Fig. \ref{fig:catvec}, it can incorporate semantic biases
from category ``Love advice'' into the words (e.g. ``Will'',
``distance'', ``relationship'', ``ruin'', ``love'' and so on) in the
question in ``Love advice''.  Thus, it can well apply the biases from category
``Love advice'' to the words (e.g. ``distance'' and ``relationship'') if
they specifically appear in ``Love advice''. On the other hand,
words that appear in several categories (e.g. ``will'') are biased with
several categories and thus will not be emphasized.

\subsection{Neural Answer Construction Model}

Here, we explain our model.  We first explain our approach and then the
algorithm.

\vspace{-2mm}
\paragraph{Approach}

It takes the following three approaches:

\begin{itemize}
\setlength{\parskip}{0.09cm} 
\setlength{\itemsep}{0.09cm}
 \item {\bf Design the abstract scenario for the answer:} The answer is
       constructed according to the order of the sentence types defined by
       the designer. For example, there are the sentence types such as
       sentence that states sympathy with the question, sentence that
       states a conclusion to the question, sentence that supplements
       the conclusion, and sentence that states encouragement to the
       questioner. This is inspired by the automated web service
       composition framework (\cite{Rao:2004:SAW:2158351.2158360}) where
       the requester should build an abstract process before the web
       service composition planning starts. In our setting, the process
       is the scenario of answer and the service is the sentence in the
       scenario. Thus, our method can construct an answer by binding
       concrete sentences to fit the scenario.

       For example, the scenario for love advice can be designed as
       follows: it begins with a sympathy sentence (e.g. ``You are
       struggling too.''), next it states a conclusion sentence (e.g. ``I
       think you should make a declaration of love to her as soon as
       possible.''), then it supplements the conclusion by a supplemental
       sentence (e.g. ``If you are too late, she maybe fall in love with
       someone else.''), and finally it ends with an encouragement
       sentence (e.g. ``Good Luck!'').


 \item {\bf Joint neural network to learn sentence selection and
       combination:} Our model computes {\em the combination optimization}
       among sentences that may include the answer as well as {\em the
       closeness} between question and sentences within a single neural
       network.
       This improves answer sentence selection; our
       model can avoid the cases in which the combination of sentences
		are
       not good enough though the scores of closeness between the question
       and each sentence are high.
       It also can let the parameter tuning simpler than the model that
       separates the network for sentence selection and that for sentence
       combination.
       The image of this neural-network is depicted in
       Fig. \ref{fig:1}-(b). Here, it learns the closeness between
       sentence ``Will distance relationship ruin love?'' and ``Distance
       cannot ruin true love'', the closeness between ``Will distance
       relationship ruin love?'' and ``Distance certainly tests your
       love.'', and the combination between ``Distance cannot ruin true
       love' and ``Distance certainly tests your love.''.
 \item {\bf Attention mechanism to improve the combination of sentences
       :} Our method extracts important topics in the conclusion
       sentence and emphasizes those topics in the supplemental sentence
       in the training phase; this is inspired by (\cite{TanSXZ16}) who
       utilizes an attention mechanism to generate the answer
       representation following the question context.  As a result, it
       can combine conclusions with the supplements following the
       contexts written in the conclusion sentences.  This makes the
       story in the created answers very natural. In
       Fig. \ref{fig:1}-(b), our attention mechanism extracts important
       topics (e.g. topic that represents ``distance'') in the conclusion
       sentence ``Distance cannot ruin true love'' and emphasizes those
       topics in computing the representation of the supplement sentence
       ``Distance certainly tests your love.''.
 \end{itemize}

\vspace{-2mm}
\paragraph{Procedure}

\begin{small}
\begin{algorithm}[t]
 \caption{A neural answer construction model}
 \label{alg:Learning}
 \begin{algorithmic}[1]
   \REQUIRE Pairs of question, conclusion, and supplement, \{($q$, $a_c$, and $a_s$)\}.
   \ENSURE Parameters set by the algorithm.
  \FOR{$n = 1$, $n{+}{+}$, while $n < N$}
   \FOR{each pair $(q, a_c, a_s)$}
      \STATE  Computes ${{\bf{o}}^c_q}$ and ${{\bf{o}}_c}$ by biLSTMs
  and max pooling.
      \STATE  Computes ${{\bf{o}}^s_q}$ by  biLSTM and max pooling.
      \FOR{each $t$-th hidden vector for supplement}  
       \STATE  Computes $\tilde{{\bf{h}}}_s{(t)}$ by Eq. (1).
      \ENDFOR
      \STATE  Computes ${{\bf{o}}_s}$ by max pooling.
      \STATE  Computes $\cal{L}$ by Eq. (2).
  \ENDFOR
    \ENDFOR
 \end{algorithmic}  
\end{algorithm}
\end{small}

The core part of the answer is usually the conclusion sentence and its
supplemental sentence. Thus, for simplicity, we here explain the
procedure of our model in selecting and combining the above two types of
sentences.  As the reader can imagine, it can easily be applied to four
sentence types. Actually, our love advice service by AI in oshiete-goo
was implemented for four types of sentences, sympathy, conclusion,
supplement, and encouragement (see Evaluation section).
The model is illustrated in Fig. \ref{fig:1}-(b) in which the input pair
is $(q, a_c, a_s)$ where $q$ is the question, $a_c$ is a candidate
conclusion sentence, and $a_s$ is a candidate supplemental sentence.
The word embeddings (WEs) for words in $q$, $a_c$, and $a_s$ are
extracted in the way described in the previous subsection.  The procedure of
our model is as follows (please see the Algorithm \ref{alg:Learning} also.):


(1) It iterates the following procedures  (2) to (7) $N$ times (line 1
in the algorithm).

(2) It picks up each pair ($q$, $a_c$, and $a_s$) in the dataset (line 2
in the algorithm).


In the following steps (3) and (4), the same biLSTM is applied to
both $q$ and $a_c$ to compute the closeness between $q$ and
$a_c$. Similarly, the same biLSTM is applied to both $q$ and
$a_s$. However, the biLSTM for computing closeness between $q$ and $a_c$
differs from that between $q$ and $a_s$ since $a_c$ and $a_s$ have
different characteristics.

(3) It separately applies a biLSTM over the two sequences of WEs, $q$
and $a_c$, and computes the max pooling over the $t$-th hidden vector for
question ${\bf{h}}^c_q{(t)}$ and that for conclusion
${\bf{h}}_c{(t)}$. As a result, it acquires the question embedding,
${{\bf{o}}^c_q}$ and the conclusion embedding, ${{\bf{o}}_c}$ (line 3
in the algorithm).

(4) It also separately applies a biLSTM over the two sequences of WEs,
$q$ and $a_s$, and computes the max pooling over the $t$-th hidden vector
for question ${\bf{h}}^s_q{(t)}$ to acquire the question embedding,
${{\bf{o}}^s_q}$ (line 4
in the algorithm). ${{\bf{o}}^s_q}$ is different from ${{\bf{o}}^c_q}$
since our method does not share the sub-network used for computing closeness
between $q$ and $a_c$ and that between $q$ and $a_s$ as described above.

(5) It applies the attention mechanism from conclusion to supplement.
      Specifically, given the output vector of biLSTM on the
      supplemental side at time step $t$, ${\bf{h}}_s{(t)}$, and the
      conclusion embedding, ${{\bf{o}}_c}$, the updated vector
      $\tilde{{\bf{h}}}_s{(t)}$ for each conclusion token is formulated
      as below (line 6
in the algorithm):

\vspace{-2mm}
\begin{eqnarray}
{{\bf{m}}_{s,c}(t)} &=& \tanh({\bf{W}}_{sm} {\bf{h}}_{s}(t) + {\bf{W}}_{cm} {\bf{o}}_{c} ) \\
{{\bf{s}}_{s,c}(t)} &=& \exp({{\bf{w}}_{mb}}^{\mathrm{T}}
 {\bf{m}}_{s,c}(t)) \nonumber \\
{\tilde{\bf{h}}_{s}(t)} &=& {\bf{h}}_{s}(t)  {\bf{s}}_{s,c}(t)  \nonumber
\end{eqnarray}
${{\bf{W}}_{sm}}$, ${{\bf{W}}_{cm}}$, and ${{\bf{w}}_{mb}}$ are
attention parameters. Conceptually, the attention mechanism gives more
      weights on  words that include important topics in the 
      conclusion sentence.

(6) It computes the max pooling over ${\tilde{\bf{h}}_{s}(t)}$ and
      acquires the supplemental embedding, ${{\bf{o}}_s}$ (line 8
in the algorithm).

(7) It computes the closeness between question and conclusion and that
      between question and supplement as well as the optimization
      combination between conclusion and supplement. The training
      objective is given as (line 9
in the algorithm):

\vspace{-2mm}      
\begin{eqnarray}
 {\cal{L}}  &=&\max\{0,
  M\!-\!(\cos({\bf{o}}_q, [{\bf{o}}^{+}_c, {\bf{o}}^{+}_s]) \!-\!
  \cos({\bf{o}}_q , [{\bf{o}}^{+}_c,
{\bf{o}}^{-}_s]))\} \\ \nonumber
{\ }&+&\max\{0, M\!-\!(\cos({\bf{o}}_q,[{\bf{o}}^{+}_{c}, {\bf{o}}^{+}_{s}]) \!-\! \cos({\bf{o}}_q,[{\bf{o}}_c^{-},
{\bf{o}}_s^{+}]))\} \\ \nonumber
{\ }&+&\max\{0, (1+k) M-\!(\cos({\bf{o}}_q,[{\bf{o}}^{+}_c, {\bf{o}}^{+}_s])\! -\! \cos({\bf{o}}_q,[{\bf{o}}_c^{-},
 {\bf{o}}_s^{-}]))\}\\ \nonumber
{\ }&+&\max\{0, 
M\!-\!(\cos({\bf{o}}_q,[{\bf{o}}^{+}_c, {\bf{o}}^{-}_s]) \!-\!
\cos({\bf{o}}_q, [{\bf{o}}^{-}_c,
{\bf{o}}^{-}_s]))\} \\ \nonumber
{\ }&+&\max\{0, 
M\!-\!(\cos({\bf{o}}_q,[{\bf{o}}^{-}_c, {\bf{o}}^{+}_s]) \!-\!
\cos({\bf{o}}_q, [{\bf{o}}^{-}_c,
{\bf{o}}^{-}_s]))\} \nonumber
\end{eqnarray}
where $[{\bf{y}}, {\bf{z}}]$ is the concatenation of two vectors,
${\bf{y}}$ and ${\bf{z}}$, ${\bf{o}}_q$ is $[{\bf{o}}^c_q,
{\bf{o}}_q^s]$, ${\bf{o}}^{+}$ is an output vector for a ground truth
answer, and ${\bf{o}}^{-}$ is that for an incorrect answer randomly
chosen from the entire answer space.  In the above equation, the first
(or second) term presents the loss that occurs when both
question-conclusion pair (q-c) and question-supplemental pair (q-s) are
correct while q-c (or q-s) is correct but q-s (or q-c) is incorrect.
%
%
      The third term computes the loss that occurs when both q-c and q-s are
      correct while both q-c and q-s are incorrect.
      The fourth (or fifth) term computes the loss that occurs when q-c (or
      q-s) is correct but q-s (or q-c) is incorrect while both q-c and
      q-s are incorrect.
%
%
$M$ is constant margin and $k \ (0 \! < k \! < 1)$ is a parameter
      controlling the margin.  Thus, the resulting margin for the third term is larger than
      those for other terms. 
%
In this way, by considering the case when either conclutions or
       supplements are incorrect or not, this equation optimizes the
       combinations among conclusion and supplement. In addition, it can
       take the closeness between question and conclusion (or
       supplement) in consideration by cosine similarity.


The parameter sets $\{{\bf{W}}_i,\! {\bf{W}}_f,\! {\bf{W}}_o,\!
{\bf{W}}_c,\!  {\bf{U}}_i,\! {\bf{U}}_f,\! {\bf{U}}_o,\! {\bf{U}}_c,
\!{\bf{b}}_i, \- \! {\bf{b}}_f,\!  {\bf{b}}_o,\! {\bf{b}}_c\}_c$ for
question-conclusion matching, $\{{\bf{W}}_i, \!{\bf{W}}_f,\- \!{\bf{W}}_o,
\!{\bf{W}}_c, 
\!{\bf{U}}_i,\!  {\bf{U}}_f, \!{\bf{U}}_o, \!{\bf{U}}_c,
\!{\bf{b}}_i, \!{\bf{b}}_f, \!{\bf{b}}_o,\!  {\bf{b}}_c\}_s$ for
question-supplement matching, and $\{{{\bf{W}}_{sm}},\!
{{\bf{W}}_{cm}},\! {{\bf{w}}_{mb}}\}$ for conclusion-supplement
attention are trained during the iterations. After the model is trained,
our method uses $\cos({\bf{o}}_q, [{\bf{o}}_c, {\bf{o}}_s])$ to score
the input ($q$, $a_c$, $a_s$) pair and constructs an answer that has a 
conclusion and its supplement.


%
%

{\tabcolsep=1.7mm
\begin{table}[t]
\begin{center}
\doublerulesep=1mm
\caption{Comparison of AP for answer selection.}
\scriptsize
{\tabcolsep = 1.0mm
\begin{tabular}{c|cccccc}
 &{\it QA-LSTM}  &   {\it Attentive-LSTM} &   {\it Semantic-LSTM} &
  {\it Construction} & {\it Our method}     \\ \hline
$K\!\!=\!\!1$& 0.8472       & 0.8196   & 0.8499 & 0.8816 & {\bf{\em 0.8846}}     \\\hline
$K\!\!=\!\!3$& 0.8649     &   0.844566   & 0.8734 & 0.8884 & 0.8909     \\\hline
$K\!\!=\!\!5$& 0.8653       &  0.8418   &  0.8712  & 0.8827 & 0.8845   \\\hline
$K\!\!=\!\!10$& 0.8603       &  0.8358   & 0.8658 & 0.8618 & 0.8647      \\
\end{tabular}
}
\label{tab:result}
\vspace{-4mm}
\end{center}
\normalsize
\end{table}
}

{\tabcolsep=1.7mm
\begin{table}[t]
\begin{center}
\doublerulesep=1mm
\caption{Comparison of AP for answer construction.}
\scriptsize
{\tabcolsep = 1.0mm
\begin{tabular}{c|cccccc}
 &{\it QA-LSTM}  &  {\it Attentive-LSTM} &   {\it Semantic-LSTM} &
  {\it Construction} & {\it Our method}     \\ \hline
$K\!\!=\!\!1$&     0.3262    & 0.3235    & 0.3664 & 0.3813 & {\bf{\em 0.3901}}     \\\hline
$K\!\!=\!\!3$&  0.3753   &   0.3694   & 0.4078 & 0.5278& 0.5308     \\\hline
$K\!\!=\!\!5$&  0.3813    &  0.3758   &  0.4133  & 0.5196  & 0.5271   \\\hline
$K\!\!=\!\!10$&  0.3827    &  0.3777   & 0.4151 & 0.4838& 0.4763   \\
\end{tabular}
}
\label{tab:result2}
\vspace{-4mm}
\end{center}
\normalsize
\end{table}
}

{\tabcolsep=1.7mm
\begin{table}[t]
\begin{center}
\scriptsize \doublerulesep=1mm \caption{Comparison of human evaluation results.}
{\tabcolsep = 1.0mm
 \begin{tabular}{c|c|c|c|c|c|c|c|c|c}
   \hline
 \multirow{2}{*}{{\it QA-LSTM}} &(1) & (2)  & (3) & (4)  & \multirow{2}{*}{{\em Our method}} &
 (1) & (2) & (3) & (4)\\ \cline{2-5}\cline{7-10}
 &  57 &  66 & 41 & 36 & & {\bf{\em 116}} & 51  & 13 & 20 \\ \hline
 \end{tabular}
}
\label{tab:result3}
\end{center}
\vspace{-4mm}
\normalsize
\end{table}
}

{\tabcolsep=1.7mm
\begin{table*}[t]
\begin{center}
\scriptsize \doublerulesep=1mm \caption{Examples of answers created by
 {\it QA-LSTM} and those by {\em Our method}.}
{\tabcolsep = 1.0mm
\begin{tabular}{p{4.5cm}|p{4.5cm}|p{4.5cm}}
Questions & {Answers created by {\it QA-LSTM}} & Answers created by {\em Our method} \\
\hline

I'm afraid to confess my love to her, what should I do?
& You should wait until you feel excited. If you interact with her
indifferently, it will be difficult to develop any relation with her.
& It is better to concentrate on how to confess your love to her.  I understand you
	 are struggling since you love her very much.	 
	 \\ \hline
	 A guy I like says to me ``I like you at home'' kiddingly. It
	 may be the ordinary gentleness. Some hope?
	 & You don't have to test his love immediately. Unless he likes
	 you, he would not have gone to see a movie with you.
	 & Yes, there is some hope. You can understand his reaction more
	 easily if your understanding of each other is deeper.
	 \\ \hline
	 I seldom meet an interesting person. I worry about how to become close to
	 him. Should I approach to him positively?
	 & Try to select your words correctly. Unless you confess your
	 love to him, it is difficult to convey your emotion to him.
	 & You should confess your love to him. Unless you
	 confess your love to him, it is difficult to convey your
	 emotion to him.
	 \\ \hline \end{tabular} }

	 \label{tab:result4} \end{center} \vspace{-4mm}
	 \normalsize \end{table*} }

\section{Evaluation}

We used our method to select or construct answers to the questions
stored in ``Love advice'' category.

\subsection{Dataset}

We evaluated our method using a dataset stored in Japanese online QA
service Oshiete-goo. First, the word embeddings were built by using
189,511 questions and their 771,956 answers stored in 16 categories
including ``Love Advice'', ``Traveling'', and ``Health Care''. 6,250
title tokens were extracted from the titles.  Then, we evaluated answer
selection and construction tasks by using a corpus containing about
5,000 question-conclusion-supplement sentences. Conclusions and
supplement sentences were extracted by human experts from answers. The
readers could use sentence extraction methods
(\cite{Schmidt:2014:DST:2637748.2638409,Zhang:2008:STB:1599081.1599218,Nishikawa:2010:OSI:1944566.1944671,Chen:2010:OEP:2898607.2898768})
or neural conversation models like (\cite{DBLP:journals/corr/VinyalsL15})
to semi-automatically extract/generate those sentences.
%

\subsection{Compared methods}\label{sec:comparedMethods}

We compared the accuracy of the following five methods:

\vspace{-1mm}
 \begin{itemize}
  \setlength{\parskip}{0.03cm} 
\setlength{\itemsep}{0.03cm}
\item {\em QA-LSTM}  proposed by (\cite{TanXZ15}).

      
\item {\em Attentive LSTM}:  introduces an attention mechanism from
      question to answer and is evaluated as the current best answer
      selection method \cite{TanSXZ16}.

\item {\em Semantic LSTM}: performs answer selection by using
      our word embeddings biased with semantics.

\item {\em Construction}: performs our proposed answer construction
      without attention mechanism.

\item {\em Our method}: performs our answer construction with attention
      mechanism from conclusion to supplement.

 \end{itemize}

\subsection{Methodology and parameter setup}\label{sec:methodology}

We randomly divided the dataset into two halves, training dataset and
predicted one, and conducted two-fold cross validation. Results shown
later are the average values.


Both for answer selection and construction, we used Average Precision (AP)
against the top-K ranked answers in the results because we consider that
the most highly ranked answers are important for users.  If the number
of ranked items is $K$, the number of correct answers among the top-j
ranked items $N_j$, and the number of all correct answers (paired with
the questions) $D$, AP is defined as follows:

\vspace{-2mm}
\begin{equation*} 
 AP = \frac{1}{D}
  {\sum_{1 \le j \le K}{\frac{N_j}{j}}}
\label{eq:map}
\end{equation*}
For answer construction, we checked whether each method could recreate the
original answers. As the reader easily can understand, this is a much more
difficult task than answer selection and thus the values of AP will be 
smaller than the results for answer selection. 

We tried word vectors and qa vectors of different sizes, and finally set
the word vector size to $300$ and the LSTM output vectors for biLSTMs to
$50 \times 2$.
We also tried different margins in the hinge loss function, and fixed the
margin, $M$, to $0.2$ and $k$ to $1.0$. The iteration count $N$ was set to
$20$.
For our method, the embeddings for questions, those for conclusions, and
those for supplements were pretrained by {\em Semantic LSTM} before
answer construction since this enhances the overall accuracy.

We did not use attention mechanism from question to answer for {{\em
Semantic LSTM}}, {{\em Construction}} and {{\em Our method}}. This is
because, as we present in the results subsection, the lengths of
questions are much longer than those of answer sentences, and thus the
attention mechanism from question to answer became noise for sentence
selection.

\subsection{Results}\label{sec:results}

We now present the results of the evaluations.

\vspace{-4mm}
\paragraph{Answer Selection}

We first compare the accuracy of methods for answer selection. The
results are shown in Table \ref{tab:result}.  {\it QA-LSTM} and {\it
Attentive LSTM} are worse than {\it Semantic-LSTM}.  This indicates that
{\em Semantic-LSTM} can incorporate semantic information
(titles/categories) into word embeddings; it can emphasize words
according to the context they appeared and thus the matching accuracy
between question vector and conclusion (supplement) vector was improved.
{\it Attentive LSTM} is worse than {\it QA-LSTM} as described above.
{\em Construction} and {\em Our method} are better
than {\it Semantic-LSTM}.  This is because they can avoid the
combinations of sentences that are not good enough even though the
scores of closeness between questions and sentences are high. This
implies that, if the combination is not good, the selection of answer
sentences also tends to be erroneous.  Finally, {\it Our method}, which
provides sophisticated selection/combination strategies, yielded higher
accuracy than the other methods. It achieved 4.4\% higher accuracy than
{\it QA-LSTM} ({\em QA-LSTM} marked 0.8472 while {\em Our method} marked
0.8846.).

\vspace{-4mm}
\paragraph{Answer Construction}

We then compared the accuracy of the methods for answer
construction. Especially for the answer construction task, the top-1 result
is most important since many QA applications show only the top-1
answer. The results are shown in Table \ref{tab:result2}. There is no
answer construction mechanism in {\it QA-LSTM}, {\it Attentive-LSTM}, and
{\it Semantic-LSTM}. Thus we simply merge the conclusion and supplement,
each of which has the highest similarity with the question by each
method.  {\it QA-LSTM} and {\it Attentive LSTM} are much worse than {\it
Semantic-LSTM}. This is because the sentences output by {\it
Semantic-LSTM} are selected by utilizing the words that are emphasized
for a context for ``Love advice'' (i.e. category and titles).  {\em
Construction} is better than {\it Semantic-LSTM} since it
simultaneously learns the optimum combination of sentences as well as
the closeness between the question and sentences. Finally, {\em Our
method} is better than {\em Construction}.  This is because it well
employs the attention mechanism to link conclusion and supplement
sentences and thus the combinations of the sentences are more natural
than those of {\em Construction}.  {\em Our method} achieved 20\% higher
accuracy than {\it QA-LSTM} ({\em QA-LSTM} marked 0.3262 while {\em Our
method} marked 0.3901.).


The computation time for {\it our method} was less than two hours.  All
experiments were performed on
NVIDIA TITAN X/Tesla M40 GPUs, and all methods were implemented by
Python in the Chainer framework.  Thus, our method well suits real
applications. In fact, it is already being used in the love advice
service of Oshiete goo \footnote{http://oshiete.goo.ne.jp/ai}.

\vspace{-4mm}
\paragraph{Human evaluation}

The outputs of {\it QA-LSTM} and {\em Our method} were judged by two
human experts. The experts entered the questions, which were not
included in our evaluation datasets, to the AI system and rated the
created answers based on the following scale: (1) the conclusion and
supplement sentences as well as their combination were good, (2) the
sentences were good in isolation but their combination was not good, (3)
One of the selections (conclusion or supplement) was good but their
combination was not good, and (4) both sentences and their combination
were not good.  The answers were judged as good if they satisfied the
following two points: (A) the contents of answer sentences correspond to
the question. (B) the story between conclusion and supplement is
natural.

The results are shown in Table \ref{tab:result3}. Table
\ref{tab:result4} presents examples of the questions and answers
constructed (they were originally Japanese and translated into English for
readability. The questions are summarized since the original ones were
very long.). The readers can also see Japanese answers 
from our service URL presented the above.
Those results indicate that the experts were much more satisfied with
the outputs of {\em Our method} than those by {\it QA-LSTM}; 58 \% of
the answers created by {\em Our method} were classified as (1). This
is because, as can be see in Table \ref{tab:result4}, {\em Our method}
can naturally combine the sentences as well as select sentences that
match the question. It well coped with the questions that were somewhat
different from those stored in the evaluation dataset.

Actually, when the public used our love advice service (\cite{article,gtc,oshieru}), it was
surprising to find that the 455 answers created by the AI whose name is
oshi-el (uses {\em Our method}) were judged as {\em Good answers} by
users from among the 1,492 questions entered from September 6th to
November 5th\footnote{This service started on September 6th, 2016.}.
The rate of getting Good answers by oshi-el is twice that of the average
human user in oshiete-goo when we focus on users who answered more than
100 questions in love advice category.  Thus, we think this is a good
result.

\section{Conclusion}\label{sec:conclusion}


This is the first study that create answers for non-factoid
questions.  Our method incorporates the biases of semantics behind
questions into word embeddings to improve the accuracy of answer
selection. It then simultaneously learns the optimum combination of
answer sentences as well as the closeness between questions and
sentences.  Our evaluation shows that our method achieves 20 \% higher
accuracy in answer construction than the method based on the current best
answer selection method.
Our model presents an important direction for future studies on answer
generation. Since the sentences themselves in the answer are short, they
can be generated by neural conversation models as we have done recently
(\cite{nakatsuji2019conclusionsupplement}); this means that our model can be
extended to generate complete answers once the abstract scenario is
made.

\bibliography{ijcai13}
\bibliographystyle{iclr2017_conference}

\end{document}